\documentclass[runningheads]{llncs}
\usepackage[backend=bibtex,style=numeric-comp,sorting=none]{biblatex} 
\usepackage{amssymb}
\usepackage{multirow}
\usepackage{graphicx}
\usepackage{hyperref}
\usepackage{authblk}
\usepackage{wrapfig}

\newcommand{\implies}{\Rightarrow}
\newcommand{\f}[1]{\mbox{\sf #1}} 
\newcommand{\fv}[1]{\mbox{\textsf{\textsl{#1}}}}

\newcommand{\fs}[1]{\mbox{\scriptsize\sf #1}} 
\newcommand{\ft}[1]{\mbox{\tiny\sf #1}}

\title{Optimising Rolling Stock Planning including Maintenance
with Constraint Programming and Quantum Annealing%
\thanks{The presented work was partially funded by the German Federal 
    Ministry for Economic Affairs and Climate Action within the project ``PlanQK'' 
    (BMWK, funding number 01MK20005).}} 
    
\titlerunning{Optimising Rolling Stock Planning including Maintenance with CP and QA} 

\author{Patricia Bickert\inst{1}\orcidID{0000-0003-0737-7604}
\footnote{The authors are listed in alphabetical order.}
\and Cristian Grozea\inst{2}\orcidID{0000-0001-6393-1919}
\and Ronny Hans\inst{1}   
\and Matthias Koch\inst{1} \and Christina Riehn\inst{1} 
\and Armin Wolf\inst{2}\orcidID{0000-0003-3940-0792}}

\authorrunning{P. Bickert, C. Grozea, R. Hans, M. Koch, C. Riehn and A. Wolf} 

\institute{DB Systel GmbH, Jürgen-Ponto-Platz 1, 60329 Frankfurt am Main, Germany\\
     \email{firstname dot lastname @deutschebahn.com}
    \and Fraunhofer FOKUS, Kaiserin-Augusta-Allee 31, 10589 Berlin, Germany\\
     \email{firstname dot lastname @fokus.fraunhofer.de}
		 }

\begin{document}

\maketitle

\begin{abstract}
We propose and compare Constraint Programming (CP) and Quantum Annealing (QA) 
approaches for rolling stock assignment optimisation considering necessary maintenance 
tasks. In the CP approach, we model the problem with an \emph{alldifferent} constraint, 
extensions of the \emph{element} constraint, and logical implications, among others. 
For the QA approach, we develop a quadratic unconstrained binary optimisation (QUBO) 
model. For evaluation, we use data sets based on real data from Deutsche Bahn and run 
the QA approach on real quantum computers from D-Wave. Classical computers are used 
to evaluate the CP approach as well as tabu search for the QUBO model. At the current 
development stage of the physical quantum annealers, we find that both approaches tend 
to produce comparable results. 

%\keywords{Constraint-Based Planning
%\and Maintenance 
%\and Quadratic Unconstrained Binary Optimisation
%\and Quantum and Simulated Annealing 
%\and Rolling Stock Optimisation} 
\end{abstract}

\section{Introduction and Motivation}

Every day, 40,000 trains travel on the Deutsche Bahn rail network, heading to 5,700 
stations. DB Fernverkehr AG, a subsidiary company of Deutsche Bahn, provides 315 Intercity-Express (ICE) trains and 
carries about 220.000 passengers between 140 ICE train stations every day in 
2020.\footnote{https://www.deutschebahn.com/de/konzern/konzernprofil/zahlen\_fakten}
Although mathematical optimisation is a key opportunity to improve the economic viability 
of railroad companies, the problem sizes are tremendous. 
The requirements to guarantee a safe operation, e.g. periodic and non-periodic maintenance 
constraints, increase the complexity even further. Furthermore, railroad companies must be 
able to respond quickly to disruptions, e.g., technical fault on a train, to ensure an 
operation in accordance with the timetable. Many real timetable-related assignment 
problems are NP-hard, which makes short-term planning very challenging.

The acceleration promised by quantum computation could be a game changer for logistics 
companies. Quantum-based  optimisation of industrial problems became an active field of 
research, even more after Google claimed quantum supremacy in 
2019~\cite{arute2019quantum}. One of the most promising devices for this now is the 
D-Wave annealer, a special purpose optimisation machine. On the other side, 
Constraint Programming (CP) is a well established programming paradigm for solving
combinatorial search and optimisation 
problems~\cite{rossiHandbookConstraintProgramming2006}.

The paper is structured as follows: In the following section, we discuss related 
approaches based on a literature review. Subsequently, we present and explain the rolling 
stock planning problem in detail. Then, in Section~\ref{sec:CP}, we present our CP-based 
solution. Afterwards, in Section~\ref{sec:QC}, we describe our quantum computing approach. 
In Section~\ref{sec:Exams}, we explain the data we use for the evaluation and present the 
results of our two approaches. In Section~\ref{sec:discuss}, we discuss the results. The 
paper closes with conclusions and future research directions.

\section{Related Work}

A comparative survey on research activities concerning optimised rolling stock 
assignment and maintenance can be found in~\cite{laiOptimizingRollingStock2015}. 
Similar to our case, there, the focus lies on passenger transportation, where the schedule
of the trips is fixed in advance. We consider rolling stock rotation 
maximising the number of operated trips, minimising empty runs, and performing maintenance
tasks, as in~\cite{giaccoRollingStockRostering2014}. Here, as in our approach, a 
pre-processing is performed to determine feasible sequences of train services 
possibly including empty runs and maintenance tasks. 
In~\cite{haahrSimultaneouslyRecoveringRolling2015}, rolling stock rescheduling is 
considered together with depot re-planning in order to handle short-term 
disruptions in railway traffic. There, a specialised branch-and-price-and-cut 
approach is used to handle such problems extending previous 
work~\cite{lusbyBranchandPriceAlgorithmRailway2017}. 
In~\cite{giaccoRollingStockRostering2014,% 
borndorferIntegratedOptimizationRolling2016,%
cacchianiModelsAlgorithmsCombinatorial2009,frischMixedIntegerLinear2019,%
katoMixedIntegerLinear2019,laiOptimizingRollingStock2015,%
reutherMathematicalOptimizationRolling2017,zhongMixedIntegerLinear2020}, 
Mixed Integer Programming (MIP) is applied to model and solve rolling stock and 
related problems --- like locomotive scheduling --- which seems to be a common approach 
for such problems. 
In~\cite{tomiyamaDevelopmentReactiveScheduling2018} constraint propagation together with 
depth-first search based on backtracking was used to perform
reactive scheduling for rolling stock operations. Similarly, 
in~\cite{haraborRailCapacityModelling2016} CP modelling and solving was 
applied for capacity maximisation of an Australian railway system transporting 
coal from mines to harbours. There, CP was able to solve this
problem finding coal train schedules which are close to 
optimistic upper capacity bounds computed analytically. Both CP approaches
encourage us to consider CP for our rolling stock optimisation problem.

A routing problem concerning railway carriages in a railway network, which is similar 
to rolling stock problems, is addressed 
in~\cite{bruckerRoutingRailwayCarriages1999}. There, a local search approach, namely 
Simulated Annealing (SA), is applied to solve the problem, because an 
integer programming approach fails. Quantum optimisation based on a QUBO model has been 
employed by~\cite{dominoQuadraticHigherorderUnconstrained2022}, where delay and conflict 
management on a single-track railway line is considered. A similiar QUBO
approach has also been used, e.g., for flight assignment 
tasks~\cite{10.1007/978-3-030-14082-3_9,vikstalApplyingQuantumApproximate2020} or traffic 
flow optimisation~\cite{10.1145/3412451.3428500}.

\section{Rolling Stock Planning with Maintenance}  
\label{sec:model}
Here, all parameters of rolling stock planning problems are
presented, i.e. the stations and the trains performing trips between them. 
Here a \emph{trip} is an entire end-to-end journey between two stations 
with potentially many stops along the way performed by one train.
The trains are potentially maintained between their trips -- either periodic or
non-periodic. The time resolution is in minutes [min] and distances are in kilometres
[km]. The considered scheduling horizon is defined by an ordered set~$H \subset 
\mathbb{N}$ of time points in minutes. Typically~$H = \{0, \ldots, e-1\}$ where 
$0$ is the canonical begin of the scheduling horizon and $e-1$ its end, e.g. $H = \{0, 
\ldots, 1439\}$ represents one day.

\subsection{Data Objects, Attributes, Constraints and Objectives}

{\bf Stations:} let $B =\{b_0, \ldots, b_{l-1}\}$ be the set of \emph{stations}.
For each pair of stations~$b, c \in B, b \ne c$ let
\begin{itemize}
    \item $\f{travelDistance}(b,c) \in \mathbb{N}^+$ and $\f{edriveDistance}(b,c) \in
        \mathbb{N}^+$ be the \emph{distances for travelling resp. for 
        driving empty} from station~$b$ to station~$c$ in [km].
    \item $\f{travelDuration}(b,c) \in \mathbb{N}^+$ and $\f{edriveDuration}(b,c) \in
        \mathbb{N}^+$be the \emph{durations for travelling resp. for 
        driving empty} from station~$b$ to station~$c$ in [min].
\end{itemize}
Obviously, we assume that $\f{travelDistance}(b,b) = \f{edriveDistance}(b,b) = 0$ and 
$\f{travelDuration}(b,b) = \f{edriveDuration}(b,b) = 0$ holds for each station~$b 
\in B$.\footnote{For the sake of simplicity we assume a homogeneous train fleet, i.e.
such that the durations for travelling between stations do not depend on the trains.}
Sometimes we abbreviate $\f{edriveDistance}(b,c)$ by $\f{edD}(b,c)$.

{\bf Trips:} 
\label{sec:trips}
let $F=\{f_0, \ldots, f_{n-1}\}$ be the set of \emph{trips}, where $n$ is the number of 
trips. For each trip~$f \in F$ let
\begin{itemize}
  \item $f.\f{departureStation} \in B$ (abbr. \f{dStation}) and $f.\f{arrivalStation} 
    \in B$  (abbr. \f{aStation}) be the \emph{departure} resp. the \emph{arrival station} 
    where the trip~$f$ starts resp. ends. 
  \item $f.\f{departureTime} \in H$ and $f.\f{arrivalTime} \in H$ be the \emph{departure}
  resp. the \emph{arrival time} when the trip~$f$ starts resp. ends. 
  \item $f.\f{travelDistance} \in \mathbb{N}^+$ be the \emph{travelling distance} of the
    trip~$f$.
  \item $f.\f{travelDuration} \in H$ be the \emph{travelling duration} of the trip~$f$.
  \item $f.\f{postProcessing} \in H$ be some \emph{post-processing} time used for 
    preparing the train after the trip~$f$.\footnote{In the considered scenarios, we
    used 120~min. for each trip, e.g. for cleaning etc.}
\end{itemize} 
It is assumed that for each trip~$f \in F$ it holds that $f.\f{departureTime} < 
f.\f{arrivalTime}$ and that $\max_{f \in F} f.\f{arrivalTime} = e-1$, i.e. the end 
of the scheduling horizon is defined by the latest arriving trip. 

{\bf Maintenance types:}
\label{sec:maint}
they are either \emph{periodic} or \emph{non-periodic}. 
Let $P$ be the set of periodic maintenance types, let $S$ be the set of non-periodic 
maintenance types. Obviously, the both sets are disjoint: $P \cap S = \emptyset$. 
Further let $W = P \cup S$ be all maintenance types. For each maintenance type~$w 
\in W$ let
\begin{itemize}
  \item $w.\f{stations} \subseteq B$ be the set of \emph{stations} where maintenance 
    tasks of type~$w$ can be performed. 
  \item $w.\f{duration} \in H$ be the \emph{duration} of the maintenance tasks of 
    type~$w$. 
  \item $w.\f{limit} \in \mathbb{N}^+$ be either the length of the maintenance interval 
    of a periodic maintenance or the threshold of a non-periodic maintenance -- both in
    [km]. Within this limit the according maintenance task~$w$ has to be performed.
\end{itemize}

{\bf Trains:} let $Z=\{z_0, \ldots, z_{m-1}\}$ be the trains potentially performing the 
trips in~$F$, where $m$ is the number of trains. For each train~$z \in Z$ let
\begin{itemize}
  \item $z.\f{initialStation} \in B$ be the \emph{initial station} from where the 
    train~$z$ starts its first trip, if the train performs any trips. 
  \item $z.\f{earliestTime} \in H$ be the \emph{earliest available time} of the 
    train~$z$. 
  \item $z.\f{initialKm}(p) \in \mathbb{N}^+$ be the initial \emph{kilometre reading} of
    the train~$z$ since the last periodic maintenance of type~$p \in P$ performed on 
    this train (cf. Sec.~\ref{sec:maint}). 
  \item $z.\f{initialKm} \in \mathbb{N}^+$ be the initial \emph{kilometre reading} of
    the train~$z$ used for non-periodic maintenance types performed on this train 
    (cf. Sec.~\ref{sec:maint}). 
\end{itemize}

{\bf Constraints and Objectives: } the rolling stock problem including maintenance 
is characterised by the  following constraints:
\begin{itemize}
    \item each train performing a trip must be available in time at the departure 
        station of the trip.
    \item all maintenance intervals -- either periodic or non-periodic -- of the trains 
        must be respected, for regular trips as well as for empty trips. 
\end{itemize}
The objective of the rolling stock problem including maintenance and idle times is to 
allocate as much as possible trips to trains, to reduce the number of empty driven 
kilometres such that the specified constraints are satisfied.

\section{Constraint-Programming-Based Solution}\label{sec:CP}

In this section a formal model of the considered rolling stock problem is presented 
which is appropriate for Constraint Programming. 

\subsection{Abstractions and Variables}\label{sec:AbsVars}

For each (real) trip~$f \in F$ we define the parameters
\begin{eqnarray*}
    \f{start}(f) = f.\f{departureTime} & \mbox{and} & 
    \f{end}(f) = f.\f{arrivalTime} \\
    \f{travelDuration}(f) = f.\f{travelDuration} & \mbox{and} & 
    \f{postProcessing}(f) = f.\f{postProcessing} \\
    \f{origination}(f) = f.\f{departureStation}  &  \mbox{and} & 
    \f{destination}(f) =  f.\f{arrivalStation} \enspace.
\end{eqnarray*}
For each train~$z \in Z$ we define a `virtual' zero-duration \emph{initial trip} $f_z$, 
where $f_{z_i} = f_{n+i}$ by convention such the all trips have unique numbers (indices), 
with 
\begin{eqnarray*}
    \f{start}(f_z) = z.\f{earliestTime}  & \mbox{and} & 
    \f{end}(f_z) = z.\f{earliestTime} \\
    \f{travelDuration}(f_z) = 0  & \mbox{and} & 
    \f{postProcessing}(f) = 0 \\
    \f{origination}(f_z)  = z.\f{initialStation}  & \mbox{and} & 
    \f{destination}(f_z) = z.\f{initialStation} \enspace,
\end{eqnarray*}
For each train~$z \in Z$ and for $k = 0 \ldots, n-1$ we define the sets of trips 
\begin{eqnarray}
  F(z)_0 & = & \{f_z\} \qquad\mbox{and} \nonumber\\
  F(z)_{k+1} & = & F(z)_k \cup \{f \in F \mid \f{ready}(f)= \min_{\stackrel{g \in F
  \land \fs{start}(g) \ge}{\max_{h \in F(z)_k} \fs{ready}(h)}} \f{ready}(g)\} 
  \enspace, \label{eqn:iterate}
\end{eqnarray}
where $\f{ready}(f) = \f{end}(f) + \f{postProcessing}(f)$ for each $f \in F \cup 
\{f_z \mid z \in Z\}$. Now for each~$z_i \in Z$ let $q(i)$ be the smallest number 
such that
  $F(z_i)_{q(i)+1} = F(z_i)_{q(i)}$
holds.  
Then $q(i)$ is a non-trivial upper bound of the number of trips performed by the 
train~$z_i$ or of the number of slots for trips for each train that maximally fit 
into the scheduling horizon~$H$. 

We would like to point out that $q(i)$ is the smallest number of steps in order to 
extend the set~$F(z)_0$ based on Equation~(\ref{eqn:iterate}) until a fixed point 
is reached. By complete induction we can prove that $F(z)_k$ contains~$k$ different
'real' trips that can be scheduled after the 'virtual' initial trip~$f_z$ of 
train~$z$ without temporal overlapping, without considering time for any 
maintenance nor any empty drives between different locations if $F(z)_k$ is a 
proper subset of $F(Z)_k+1$. If $F(z)_k = F(z)_k+1$ holds then there are at most 
$k$ 'real' trips that can be scheduled in that way after the 'virtual' initial 
trip~$f_z$ of train~$z$. Considering additional maintenance tasks or 
empty drives between the trips performed by any train~$z_i$ the number of trips 
that can be performed by this train is not greater than~$q(i)$ such that~$q(i)$ 
is a non-trivial upper bound of the number of 'real' trips the can be performed 
by train~$z_i$. Due to the fact that the sets~$F(z)_k$ are only used for the
calculation of these bounds, no solutions are lost. 

Based on~$q(i)$ we define for each train~$z_i \in Z$ a sequence of finite domain
variables~$\fv{Trip}_{i,0} = n+i$ (i.e. the 
initial trip first) and
\begin{eqnarray*} 
    \fv{Trip}_{i,1} \in \{-(iq+1), 0, \ldots, n-1\}, & \ldots, &
    \fv{Trip}_{i,q(i)} \in \{-(iq+q), 0, \ldots, n-1\} \enspace,
\end{eqnarray*} 
where $q = \max_{i \in \{0, \ldots, m-1\}} q(i)$ is the greatest number of trips for 
any train. Theses variables are presenting the indices of potential trips that will be 
performed by the train~$z_i$ according to the order of the sequence, i.e. 
$\fv{Trip}_{i,j}$ indicates the potential trip performed by train~$z_i$ in 
\emph{time slot}~$j$.
By definition $\fv{Trip}_{i,j} < 0$ indicates  that train~$z_i$ will not perform any 
`regular' (neither `real' nor `virtual') trip in slot~$j$. We have chosen unique 
negative values within the domains of these variables such they can be pairwise different
(see Equation~(\ref{eq:alldiff}) below).
In addition to these $\fv{Trip}$ variables we use further indexed variables: 
\begin{itemize}
    \item $\fv{Start}_{i,j} \in H$: the start time of the `trip' performed 
        in slot~$j$ of train~$z_i$.
    \item $\fv{End}_{i,j} \in H$: the end time of the `trip' performed in 
        slot~$j$ of train~$z_i$.
    \item $\fv{PostProcessing}_{i,j} \in H$: the post processing time of the 
        `trip' in~$j$ of~$z_i$.
    \item $\fv{Origination}_{i,j} \in \{-1, 0, \ldots, l-1\}$: the 
        origination of the `trip' in~$j$ of~$z_i$.         
    \item $\fv{Destination}_{i,j} \in \{-1, 0, \ldots, l-1\}$: the 
        destination of the `trip' in~$j$ of~$z_i$.         
    \item $\fv{OptDuration}_{i,j} \in H$: the optional duration of a 
        maintenance task performed after the `trip' in slot~$j$ of train~$z_i$.
    \item $\fv{OptLocation}_{i,j} \in \{-1, 0, \ldots, l-1\}$: either the
        optional location of a maintenance task performed after the `trip' performed in 
        slot~$j$ of train~$z_i$ or the destination of this trip.    
    \item $\fv{DurationTo}_{i,j} \in H$ and $\fv{DurationFrom}_{i,j} \in H$: the 
        durations after the `trip' in slot $j$ to resp. from an optional maintenance of
        train~$z_i$.
    \item $\fv{DistanceTo}_{i,j} \in H$ and $\fv{DistanceFrom}_{i,j} \in H$: the 
        distance after the `trip' in slot $j$ to resp. from an optional maintenance task
        of train~$z_i$.
    \item $\fv{Distance}_{i,j} \in \mathbb{N}^+$: the distance 
        of the `trip' in slot $j$ of train~$z_i$.
    \item $\fv{IsRegularTrip}_{i,j} \in \{0, 1\}$: signals whether the `trip' in $j$ 
        of~$z_i$ is `regular'.
    \item $\fv{IsMaintained}(w)_{i,j} \in \{0, 1\}$: signals whether there is 
        a maintenance task of type~$w$ after the `trip' in slot $j$ of train~$z_i$.
    \item $\fv{KmReading}(p)_{i,j} \in \mathbb{N}^+$: the kilometre reading for a 
        periodic maintenance task of type~$p$ of train~$z_i$ directly after the `trip'
        in slot $j$.
    \item $\fv{KmReading}_{i,j}$: the kilometre reading for all non-periodic 
        maintenance tasks of train~$z_i$ directly after the `trip' in slot $j$.
\end{itemize}

\subsection{Constraints}\label{sec:Constraints}

Due to the fact that each trip will be performed by at most one train all these trip
variables must have pairwise different values:%
\footnote{cf. \url{https://sofdem.github.io/gccat/gccat/Calldifferent.html}}
\begin{equation}
  \f{allDifferent}(\{\fv{Trip}_{i,j} \mid i \in \{0, \ldots, n-1\} % z_i \in Z 
  \land j \in \{1, \ldots,q(i)\}\}) \enspace. \label{eq:alldiff}
\end{equation}
Due to the fact that each sequence of trips performed by a train~$z_i$ is without gaps, 
it must hold that
\begin{equation}
  \fv{Trip}_{i,j-1} < 0 \implies \fv{Trip}_{i,j} < 0 \quad\mbox{resp.}\quad
  \fv{Trip}_{i,j} \ge 0 \implies \fv{Trip}_{i,j-1} \ge 0 \label{eq:pruneIrregularTrips}
\end{equation}
for $i=0, \ldots, n-1$ and $j=1, \ldots, q(i)$ and further 
\begin{equation}
    \fv{IsRegularTrip}_{i,j} \Leftrightarrow \fv{Trip}_{i,j} \ge 0
    \quad \mbox{for $i=0, \ldots, n-1$ and $j=0, \ldots, q(i)$.}
\end{equation}
For the set of indices~$I = \{0, \ldots, n+m-1\}$ of all trips 
additional (extended) \emph{element} 
constraints\footnote{cf.~\url{https://sofdem.github.io/gccat/gccat/Celement.html}} 
must hold on indexed variables, when $k = \fv{Trip}_{i,j}$:
\begin{eqnarray}
    \fv{Start}_{i,j} & = & \fv{Start}[\fv{Trip}_{i,j}] 
        = \left\{\begin{array}{rl}
            \f{start}(f_k) & \quad\mbox{for $k \in I$} \\
            e-1 & \quad\mbox{for $k < 0$}
    \end{array}\right.
\end{eqnarray}
and analogously $\fv{End}_{i,j}  =  \fv{End}[\fv{Trip}_{i,j}]$ where $\f{start}(f_k)$ is
replaced by~$\f{end}(f_k)$ and $\fv{PostProcessing}_{i,j}  =  
\fv{PostProcessing}[\fv{Trip}_{i,j}]$ where 
\begin{eqnarray*}
    \fv{PostProcessing}[\fv{Trip}_{i,j}]
        & = & \left\{\begin{array}{rl}
            \f{postProcessing}(f_k) & \quad\mbox{for $k \in I$} \\
            0 & \quad\mbox{for $k < 0$}
    \end{array}\right. \quad\mbox{and} \\
    \fv{Destination}_{i,j} & = & \fv{Destination}[\fv{Trip}_{i,j}] 
        = \left\{\begin{array}{rl}
            \f{destination}(f_k) & \quad\mbox{for $k \in I$} \\
            -1 & \quad\mbox{for $k < 0$}
    \end{array}\right.
\end{eqnarray*}
as well as $\fv{Origination}_{i,j}$ 
where $\f{destination}(f_k)$ is replaced by~$\f{origination}(f_k)$.

There is at most one maintenance task even after each `real' trip performed by a train, 
i.e. for each train~$z_i$ and each slot~$j$ it must hold that
\begin{eqnarray}
  \sum_{w \in W} \fv{IsMaintained}(w)_{i,j} \le 1 \enspace.
  \label{eq:atMostOneMaintenance}
\end{eqnarray}
If there is not any trip performed by train~$i$ in slot~$j$ or in slot~$j+1$ then there 
will not be any maintenance task afterwards:
\begin{eqnarray}
  \fv{Trip}_{i,j} < 0 \lor \fv{Trip}_{i,j+1} < 0 
  & \implies &  \fv{IsMaintained}(w)_{i,j} = 0 \quad\mbox{for each~$w \in W$}  
  \nonumber \\
  \mbox{resp.} \quad \fv{Trip}_{i,j} < 0 \lor \fv{Trip}_{i,j+1} < 0 
  & \implies & \sum_{w \in W} \fv{IsMaintained}(w)_{i,j} = 0 \enspace.
\end{eqnarray}
For each type of maintenance~$w \in W$, each train~$z_i \in Z$ and each time slot~$j$ 
it holds  that the optional duration and the optional locations for the maintenance of 
this type are defined if the maintenance occurs or not:
\begin{eqnarray}
    \fv{IsMaintained}(w)_{i,j} = 1 & \implies & \fv{OptDuration}_{i,j}  =  
        w.\f{duration} \nonumber \\
    & & \land~\fv{OptLocation}_{i,j} \in w.\f{stations} \label{eq:MaintStations} \\
    \sum_{w \in W} \fv{IsMaintained}(w)_{i,j} = 0 & \implies & \fv{OptDuration}_{i,j} = 
        0 % \label{eq:noMaintOne} 
        \nonumber \\ [-2ex]
    & & \land~\fv{OptLocation}_{i,j} = \fv{Destination}_{i,j} \label{eq:noMaintTwo}
\end{eqnarray}

The trips of each train are performed in linear order and there must be enough 
time for empty transition drives and optional maintenance tasks, i.e. for each 
train~$z_i \in Z$ and each time slot~$j=0, \ldots, q(i)-1$ it must hold that
\begin{eqnarray}
    \lefteqn{\fv{Start}_{i,j+1} \ge \fv{End}_{i,j}  %% + \fv{IdleTime}_{i,j} 
    + \fv{PostProcessing}_{i,j}} \nonumber \\ 
    & & ~+~ \fv{OptDuration}_{i,j} +\fv{DurationTo}_{i,j} + \fv{DurationFrom}_{i,j} 
    \enspace.
\end{eqnarray}
Here we use two-dimensional extended \emph{element} constraints:
\begin{eqnarray}
    \fv{DurationTo}_{i,j} 
    & = & \fv{EDriveDuration}[\fv{Destination}_{i,j}][\fv{OptLocation}_{i,j}] 
    \label{eq:DurationTo} \\
    \fv{DurationFrom}_{i,j} 
    & = & \fv{EDriveDuration}[\fv{OptLocation}_{i,j}][\fv{Origination}_{i,j+1}] 
        \label{eq:DurationFrom}
\end{eqnarray}
where $\fv{EDriveDuration}[\underline{~}][\underline{~}]$ represents the duration for
empty drivings, i.e. 
\begin{eqnarray}
    \fv{EDriveDuration}[r][s] 
        = \left\{\begin{array}{rl}
            \f{edriveDuration}(b_r, b_s) & \quad\mbox{for $r, s \in 0, \ldots, l-1$} \\
            0 & \quad\mbox{for $r < 0$ or $s < 0$}
    \end{array}\right. 
\end{eqnarray}
and analogously $\fv{TravelDuration}[r][s]$ where $\f{edriveDuration}(b_r, b_s)$ is 
replaced by~$\f{travelDuration}(b_r, b_s)$ representing the travelling durations between 
stations. 

The kilometre reading for all non-periodic maintenance tasks of train~$z_i$  are
defined by $\fv{KmReading}_{i,0} = z_i.\f{initialKm}$ and directly 
after the trip in slot $j$ by
\begin{eqnarray} 
  \fv{KmReading}_{i,j+1} & = & \fv{KmReading}_{i,j} + \fv{DistanceTo}_{i,j} \nonumber \\
  & & ~+~\fv{DistanceFrom}_{i,j} + \fv{Distance}_{i,j+1} \enspace.
\end{eqnarray}
Here we use two-dimensional extended \emph{element} constraints, too:
\begin{eqnarray}
    \fv{DistanceTo}_{i,j} 
    & = & \fv{EDriveDistance}[\fv{Destination}_{i,j}][\fv{OptLocation}_{i,j}] \\
    \fv{DistanceFrom}_{i,j} 
    & = & \fv{EDriveDistance}[\fv{OptLocation}_{i,j}][\fv{Origination}_{i,j+1}] \\
    \fv{Distance}_{i,j+1} 
    & = & \fv{TravelDistance}[\fv{Origination}_{i,j+1}][\fv{Destination}_{i,j+1}]
\end{eqnarray}
where $\fv{EDriveDistance}[\underline{~}][\underline{~}]$ represents the empty driving 
distances, i.e. 
\begin{eqnarray}
    \fv{EDriveDistance}[r][s] 
        = \left\{\begin{array}{rl}
            \f{eDriveDistance}(b_r, b_s) & \quad\mbox{for $r, s \in 0, \ldots, l-1$} \\
            0 & \quad\mbox{for $r < 0$ or $s < 0$}
    \end{array}\right. 
\end{eqnarray}
and analogously $\fv{TravelDistance}[r][s]$ where $\f{eDriveDistance}(b_r, b_s)$ is 
replaced by~$\f{travelDistance}(b_r, b_s)$
representing the travelling distances. 

After the `initial' trip where the kilometre reading is defined by the 
initial kilometre reading for all successive slots~$j+1$ with $j \in \{0, \ldots, 
q-1\}$ it holds that the kilometre reading is the kilometre reading of the 
previous slot~$j$ plus the distances for all trips performed in between. 
Furthermore, the limits of all non-periodic maintenance tasks must be respected, i.e. 
if for any train~$z_i$ the kilometre reading exceeds the threshold of a non-periodic 
maintenance of type~$s$ after a trip in slot~$j \ge 0$ then this maintenance must be 
performed directly before this trip:
\begin{eqnarray*}
  \fv{kmReading}_{i,j} \le s.\f{limit} \land \fv{kmReading}_{i,j+1} > 
    s.\f{limit}
  & \implies & \fv{IsMaintained}(s)_{i,j} = 1 \enspace.
\end{eqnarray*}
For the kilometre reading of a periodic maintenance~$p$ of train~$z_i$ let
\begin{eqnarray*}
    \fv{KmReading}(p)_{i,0} & = &  z_i.\f{initialKm}(p) \quad \mbox{where 
    $z_i.\f{initialKm}(p) \le p.\f{limit}$}
\end{eqnarray*}
and directly after  the trip in slot $j \ge 0$ it must hold that
\begin{eqnarray*}
    \lefteqn{\fv{KmReading}(p)_{i,j+1} = \fv{DistanceFrom}_{i,j} + \fv{Distance}_{i,j+1}}
    \nonumber \\
    & & +~ (1 - \fv{IsMaintained}(p)_{i,j}) \nonumber \cdot~(\fv{KmReading}(p)_{i,j} + 
    \fv{DistanceTo}_{i,j})
\end{eqnarray*}
and $\fv{KmReading}(p)_{i,j+1} \le p.\f{limit}$ to satisfy the maintenance limits.

\subsection{Simplifications}
If there are similar trains then there is a maximum distance~$\f{maxD}(H)$ 
that each train can drive during the scheduling horizon~$H$. 
Then for each train~$z_i \in Z$ we can decide in advance whether
there are any maintenance tasks to be performed: If 
    $\fv{KmReading}_{i,0} + \f{maxD}(H) \le s.\f{limit}$
holds for each non-periodic maintenance of type~$s$ and 
    $\fv{KmReading}(p)_{i,0} + \f{maxD}(H) \le p.\f{limit}$
holds for each periodic maintenance of type~$p$ then the 
variables~$\fv{IsMaintained}(w)_{i,j}$ as well as 
Conditions~(\ref{eq:atMostOneMaintenance})--(\ref{eq:MaintStations}) 
can be omitted and Condition (\ref{eq:noMaintTwo}) can be simplified to:
    $\fv{OptDuration}_{i,j} = 0$ and
    $\fv{OptLocation}_{i,j} = \fv{Destination}_{i,j}$.

\subsection{Objective}\label{sec:objective}

The objective of the rolling stock planning with maintenance tasks is to maximise the 
number of performed trips while minimising the empty driven kilometres. Therefore we 
maximise the weighted sum
\begin{eqnarray}
    \lefteqn{2\cdot D \sum_{i=0}^{m-1} \sum_{j=1}^{q(i)} 
    \fv{IsRegularTrip}_{i,j}} \nonumber \\
    & & - \sum_{i=0}^{m-1} \sum_{j=0}^{q(i)-1} 
    (\fv{DistanceTo}_{i,j} + \fv{DistanceFrom}_{i,j})
    \label{eqn:obj}
\end{eqnarray}
where the factor $2\cdot D = 2\cdot \max_{(b,c) \in B \times
B}\f{edriveDistance}(b,c)$
is twice the maximal distance of all empty drives assuming that
$\fv{DistanceTo}_{i,j} + 
\fv{DistanceFrom}_{i,j} < 2\cdot D$ holds for any trip~$i$ and any 
slot~$j$.\footnote{otherwise
use $2\cdot D + 1$ or $2\cdot(D + 1)$ instead.} 
This ensures that in the objective~(\ref{eqn:obj}) the first addend dominates 
the second subtrahend such that the number of performed trips is the major objective 
and empty driven kilometres is the subordinate objective, as business requirements specify.  
The choice of the factor $2 \cdot D$ is sufficient. It ensures that if a trip takes 
place, then the possibly negative contribution due to empty trips is less than 
$2 \cdot D$, i.e. for a trip the amount to the objective is always positive. 
It is greater, the shorter any empty runs. If no trip takes place in the slot~$j$ of a
train~$i$, then, by definition, $\fv{DistanceTo}_{i,j} = \fv{DistanceFrom}_{i,j} = 0$ 
holds.

The number of maintenance tasks are not explicitly minimised because a maintenance task 
is only performed if necessary and minimising the empty driven kilometres implicitly 
covers the minimisation of empty driven kilometres to and from maintenance tasks, too.

\section{Quantum-Computing-Based Solution} \label{sec:QC}

For a quantum optimisation approach, we developed a model based on quadratic
unconstrained binary optimisation (QUBO) \cite{glover2018tutorial,% 
kochenberger2014unconstrained}. The QUBO models have the advantage of being 
hardware independent, approachable both on the gate-based universal 
quantum computers with QAOA~\cite{farhi2014quantum} and on adiabatic quantum computers such as D-Wave
machines~\cite{harris2010experimental} with quantum annealing (QA).

\subsection{QUBO Model}

The heart of our QUBO model are the assignment variables, which link the 
available trains to the trips. In our model, each train is able to operate at 
most $q$ trips, with $q$ being defined beforehand.\footnote{This is a 
simplification w.r.t. the ``train-specific'' value~$q(i)$ introduced in 
Section~\ref{sec:AbsVars}.}
This leads to the binary decision variable $X[i,f,z]$, where $f \in F$ defines 
the trip, $z \in Z$ the train and $i \in \{0, \dots, q-1 \}$ corresponds to the 
time slot. If a trip $f$ is operated by train $z$ in time slot $i$, $X[i,f,z]$ then it 
is equal to $1$, otherwise $0$. We used $q=3$ for all trains.
In order to obtain a valid timetable, we have to enforce the following three constraints:
\begin{enumerate}
    \item Each train operates at most one trip in each time slot.
    \label{cons:one}
    \item Each trip is operated at most once and by a single train.
    \label{cons:two}
    \item Successive trips operated by the same train do not overlap (there is
    sufficient time for a possible necessary empty trip and the required 
    post-processing).
    \label{cons:three}
\end{enumerate}

The next step is the inclusion of maintenance. To this end, we integrate maintenance 
actions $w \in W$ in a new set of (extended) trips $F'_{f,w}$.
This new set of trips consists of duplicates of the original 
trip $f \in F$ from the timetable, but for each trip $f' \in F'_{f,w}$, a 
maintenance action $w$ is conducted before the regular trip $f$ takes place. 
In total, for each service station able to conduct a maintenance $w$ an optional 
maintenance trip is created and added to $F'_{f,w}$. 
Consequently, each trip $f' \in F'_{f,w}$ starts from one of the possible 
maintenance stations, such that
$f'.\f{maintenanceStation} \in w.\f{stations}\ (\mathrm{abbr.}\ \f{mStation})$, 
$f'.\f{departureStation} = f'.\f{maintenanceStation}$,
and its duration exceeds that of $f$ by the maintenance duration and the travel 
duration from the maintenance station to the start station of~$f$,
$f'.\f{duration}  =  f.\f{travelDuration} + w.\f{duration}
+\,  \f{eDriveDuration}(w.\f{station}, f.\f{departureStation}).
$
Also the departure time is adapted accordingly,
$ f'.\f{departureTime}  =  f.\f{departureTime} - w.\f{duration}
-\, \f{eDriveDuration}(w.\f{station}, f.\f{departureStation}).$
The integration of the maintenance actions increases the number of decision 
variables to 
$n \cdot m \cdot q \cdot \sum_{w \in W} |w.\f{stations}|$.
This approach requires Constraint~\ref{cons:two} to be modified, as at most one 
trip of $F'_{f} \cup \{f\}$ needs to be operated, where $F'_{f} = \cup_{w \in W}
F'_{f,w}$ contains all optional maintenance trips that are obtained from the
regular trip $f \in F$ for all maintenance $w \in W$. 
The set $F_{\fs{all}} =  \cup_{f \in F} (F'_{f} \cup \{f\})$ contains all trips,
with and without maintenance, where $F$ corresponds to the regular 
trips from the original timetable (cf.~Sec.~\ref{sec:trips}).

In the proposed QUBO model, the Constraints~\ref{cons:one}--\ref{cons:three} 
are implemented as penalty terms in the objective function~(Eq.~(\ref{eq:costfunction})). 
If the constraint is fulfilled, the associated term ($\fv{c1}$-$\fv{c3}$) evaluates to 
zero:
\begin{eqnarray}
  \fv{c1} & = & \sum_{i = 0}^{q-1}\sum_{z \in Z}\sum_{\stackrel{f_1,f_2 \in 
  F_{\ft{all}}}{ 
  f_1 \ne f_2}}{X[i,f_1,z]\cdot  X[i,f_2,z]}, \label{eq:atMostOne}
  \\ 
  \fv{c2} & = & \sum_{f \in F}\sum_{f_1,f_2 \in F'_{f} \cup \{f\}} \sum_{i_1,i_2 
  = 0}^{q-1}\sum_{\stackrel{z_1,z_2 \in Z}{f_1 \ne f_2 \lor i_1 \ne i_2 \lor z_1 
  \ne z_2}} \!\!\!\!\!\!\!\!
  {X[i_1,f_1,z_1]\cdot X[i_2,f_2,z_2]},
\\
\fv{c3} & = & \sum_{\stackrel{i_1,i_2 = 0}{i_1 < i_2}}^{q-1}\sum_{z \in Z} 
\sum_{\stackrel{f_1,f_2 \in F_{\ft{all}}}{\fs{overlap}(f_1, f_2)}}{X[i_1,f_1,z] 
\cdot X[i_2,f_2,z]},
\end{eqnarray}
where the condition~$\f{overlap}(f_1, f_2)$ determines if two 
trips~$f_1, f_2$ overlap:
\begin{eqnarray*}
    \lefteqn{\f{overlap}(f_1, f_2) 
    := (f_2.\f{departureTime} ~<~ f_1.\f{arrivalTime} } \\
    & & +~ f_1.\f{postProcessing} + \f{edriveDuration}(f_1.\f{arrivalStation}, 
    f_2.\f{departureStation})) \enspace.
\end{eqnarray*}

The maintenance constraints are included heuristically in our model. We distinguish two 
cases: First, when a train~$z$ is close to the maintenance limit for a 
maintenance type~$w$, an immediate maintenance action is required,
$\f{immediateAction}(w,z) := w.\f{limit} < z.\f{initialKm}_w + 500$,
which should be preferably carried out before the first trip; 500 km is a data-based 
threshold chosen such that trains with less available range, with high probability, will 
not be able to perform regular trips. Second, if no immediate maintenance is necessary, 
maintenance actions $w$ should be performed within the scheduling horizon depending on how 
close a train $z$ is to the maintenance limit at the beginning of the optimisation. To be 
cost effective, we want to delay the maintenance as long as possible. However, 
conducting the maintenance just before reaching the maintenance limit strongly 
restricts the possible solutions of the problem. Therefore, we 
integrate the need of maintenance heuristically and slowly increase the 
maintenance penalty with increasing kilometre reading of the train. To this 
end, we calculate a heuristic weight $\alpha(w,z)$ for the maintenance penalties 
discussed below and fix it to $$\alpha(w,z) := 1/\left(e^{0.002 \cdot 
(w.\fs{limit}-1300-z.\fs{initialKm}_w)}+1\right) \enspace,$$ which increases 
continuously from $0$ to $1$ with increasing $z.\f{initialKm}_w$. 
The constant 0.002 is a damping factor reducing the weight for small $z.\f{initialKm}_u$, 
and 1300 has been chosen, because it corresponds to the average driven kilometres of a 
train being in operation for three days. This way, we increase the chances that the trains 
are not too close to the limit and have enough remaining range left for a potential next 
planning period. The cost function is then extended with the following terms:
\begin{eqnarray} 
 \fv{cm1} & = & \sum_{z \in Z}\sum_{\stackrel{w \in 
 W}{\fs{immediateAction}(w,z)}}
 \alpha(w,z)\Bigg(\sum_{f \in F}\sum_{f_1 \in F'_{f,w}}{X[0,f_1,z]}-1\Bigg)^2,
 \\
 \fv{cm2} & = &  \sum_{z \in Z}\sum_{\stackrel{w \in W}{\neg 
 \fs{immediateAction}(w,z)}} \alpha(w,z)\Bigg(\sum_{i = 0}^{i<q}\sum_{f \in 
 F}\sum_{f_1 \in F'_{f,w}}{X[i,f_1,z]}-1\Bigg)^2,
 \\
 \fv{cm3} & = &  \sum_{z \in Z}\sum_{w \in W}(1-\alpha(w,z))\sum_{i = 0}^{q-1}
 \sum_{f \in F}\sum_{f_1 \in F'_{f,w}}{X[i,f_1,z]}.
\end{eqnarray}
The first two terms promote that a maintenance action is performed. The first term forces 
maintenance to take place in the first time slot, while the second term does not 
distinguish between the time slots, but maintenance should be within the optimisation 
horizon. The third term, on the other hand, tries to prevent unnecessary maintenance. 
The significance of the two contradicting optimisation goals is controlled by 
$\alpha(w,z)$. 

It important to note that while we steer indeed heuristically with the soft constraints cm1-3 the pressure towards maintenance (and simultaneously against too much maintenance), the choice whether to perform maintenance or not, its time, and its place, is still left to the optimisation for all trains, except for those very close to exceeding a maintenance interval - here we apply more pressure that the maintenance happens before the first trip. 

As discussed in Section~\ref{sec:objective}, our aim is not only to maximise the number of 
performed trips but also to minimise the empty driven kilometres. To this end, we sum up 
the empty driven kilometres by
\begin{eqnarray*}
\lefteqn{\fv{totalEmptyKM} = \sum_{z \in Z}\Bigg[\sum_{f \in F}\bigg( X[0,f,z] 
\cdot \f{edD}(z.\f{initialStation},f.\f{departureStation})}\\ 
& & + \sum_{f' \in F'_f} X[0,f',z] \cdot \big( 
\f{edD}(z.\f{initialStation},f'.\f{maintenanceStation}) \\
& &\qquad\qquad\qquad\qquad~+~
\f{edD}(f'.\f{maintenanceStation},f.\f{departureStation})\big)\bigg)\\
 & & +~\sum\limits_{i = 0}^{q-2}\sum_{\stackrel{f_1\in F_{\ft{all}}, f_2 \in 
 F}{f_1 \ne f_2}} 
 \Bigg( X[i,f_1,z] \cdot X[i+1,f_2,z] 
 \cdot\f{edD}(f_1.\f{aStation},f_2.\f{dStation})\\
 & & \quad\qquad~+~ \sum_{\stackrel{f'_2 \in F'_{f_2}}{f'_2 \ne f_1}} \bigg(
 X[i,f_1,z] \cdot X[i+1,f'_2,z] \cdot 
 \big(\f{edD}(f_1.\f{aStation},f'_2.\f{mStation}) \\
 & & \qquad\qquad\qquad\qquad~+~ 
 \f{edD}(f'_2.\f{mStation},f_2.\f{dStation})\big)\bigg)\Bigg)\Bigg]
\end{eqnarray*}
Finally, our actual optimisation goal is to maximise the number of trips 
operated by each train, which is equivalent to minimising its negative: 
\begin{eqnarray}
\fv{optimizationGoal}
& = & - \sum_{z \in Z}\sum_{f \in F_{\ft{all}}}{X[0,f,z]}\nonumber \\ 
& & -\sum_{z \in Z}\sum_{i = 0}^{q-2}\sum_{\stackrel{f_1,f_2 \in 
F_{\ft{all}}}{f_1 \ne f_2}}
{X[i,f_1,z] \cdot X[i+1,f_2,z]}. 
\end{eqnarray}
To force the optimiser to select a trip for each time slot, we reward successive trips in 
particular. Otherwise, the optimiser might select only a trip for every second time slot, 
to avoid the \fv{totalEmptyKM} penalty (e.g. $x[0,f_1,z] = 1$, $x[1,f,z] = 0\ \forall f 
\in F_{\f{all}}$,  and $x[2,f_2,z] = 1$).

The complete cost function is then given by
\begin{eqnarray}
\label{eq:costfunction}
    & & wg_{\f{reward}} \cdot \fv{optimizationGoal} + 
        wg_{\f{penalty}}\cdot(\fv{c1}+\fv{c2}+\fv{c3})  
    \\ \nonumber
    \quad & & + \ wg_{\f{maintenance}}\cdot(\fv{cm1}+\fv{cm2}+\fv{cm3})
    + wg_{\f{km}}\cdot\fv{totalEmptyKM}.
\end{eqnarray}
We use the weights $$wg_{\f{reward}} = 100, wg_{\f{km}} = 100/(2D), 
wg_{\f{penalty}} = 1000, wg_{\f{maintenance}} = 100 \enspace.$$ 
The weight $wg_{\f{km}}$ was chosen such that the ratio with $wg_{\f{reward}}$ 
is consistent with Equation~(\ref{eqn:obj}), in order to synchronise the 
objective functions of the CP and QUBO approaches.

To reduce the problem complexity and the number of necessary qubits, we identify 
those variables which can be neglected already before the optimisation. To this 
end, we consider two cases. First, we check if a trip $f \in F_{\f{all}}$ is 
feasible for a train $z$ by evaluating the following inequality
\begin{eqnarray}
f.\f{departureTime} & < & 
\f{edriveDuration}(z.\f{initialStation},f.\f{departureStation})\\ \nonumber
& & + z.\f{earliestTime}
\end{eqnarray}
If this is true, the train cannot operate the trip in time and we do not 
consider the variable for this combination, since $X[0,f,z] = 0$. 
Second, we neglect maintenance trips $f' \in F'_{j,w}$ for a train $z$, if the 
train is far away from the maintenance limit, i.e.
$z.\f{initialKm}_w + 3000 < w.\f{limit}$,
where 3000 km is a threshold chosen based the distance a  train can drive within the 
scheduling horizon, which is one day in our case.
If this expression is true, the train will not reach the maintenance limit 
within the scheduling horizon and $X[i,f',z] = 0\ 
\forall\ i\in \{0,\dots,q-1\}$.

% \newpage %% AW: might be removed - but now it looks better!

\section{Empirical Examination of Both Approaches} \label{sec:Exams}
To be able to properly compare our CP and QUBO solutions, we started from the same mathematical model of a real-world problem described in Section \ref{sec:model}. We tried to keep both approaches as similar as possible, but not identically due to technical reasons. The differences often result from the technical limitations of current quantum hardware and they are necessary to keep the number of qubits low. 
\begin{itemize}
    \item We restrict the solution to three trips per train. This make almost no difference on our real data and our 24h planning horizon, because considering the trip length and the starting times, it was barely possible for a train to operate more than three trips during the optimisation horizon. The CP solution produce for very few trains 4 trips assignments before limiting the train specific upper bound to 3. This is enforced for the CP as well as the QUBO model.
    \item
    The maintenance constraints are implemented as a soft constraints for QUBO vs. hard constraint for CP. Thus for QUBO some of those might be violated, which is addressed in postprocessing, possibly reducing the number of covered trips. Both the number before and after postprocessing are shown.    
\end{itemize}

\begin{wrapfigure}{i}{0.37\textwidth}
    \centering
    \includegraphics[width=0.37\textwidth]{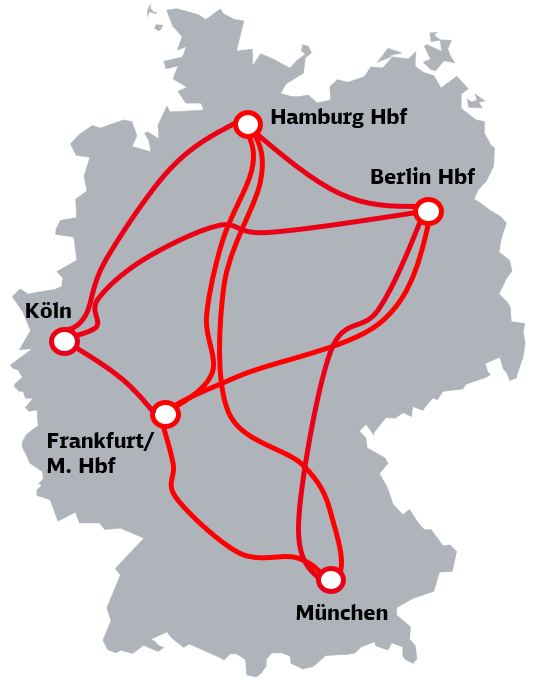}
    \caption{The examined simplified train network.} \label{fig:germanmap}
\end{wrapfigure}
For empirical examination of both approaches we have generated different-sized data sets  
considering a subset of the German rail network with trips between the major cities Berlin, 
Frankfurt, Hamburg, Munich, and Cologne, shown in Figure~\ref{fig:germanmap}.
The data sets are based on a \emph{real} train schedule for one day from Deutsche Bahn. We 
then simplified the timetable under the following aspects: first, we only consider direct 
trips between these cities and ignore intermediate stations. 
This simplification yields a timetable with 284 trips for one day.
Furthermore, we used standardised distances and travel times between the cities,
ignoring the variations that depend on the actual paths.
We consider two maintenance types, one periodic and one non-periodic maintenance, using 
realistic intervals.
Finally, we obtain different-sized subsets of our data set by varying the total number of 
trips and trains (cf.~Table~\ref{tab:results}).

We then employ our two approaches to search for good or even best trip allocations w.r.t. 
to the defined objective functions, cf.~(\ref{eqn:obj}) and (\ref{eq:costfunction}), 
respectively. 
% We use the Constraint Programming library {\tt firstCS}~\cite{wolfFirstCSNewAspects2012} 
% TODO: restore!
We use a Constraint Programming library {\tt firstCS}~\cite{wolfFirstCSNewAspects2012} 
to implement and solve our CP approach. For this purpose we implemented straightforward
filtering methods for the extended element constraints and the logical implications. 
Further, we applied monotonous \emph{branch and bound} (B\&B) using a depth-first 
tree search with a \emph{first-fail} heuristic: Trains with smaller numbers of 
potential trips or the same number of potential trips but with a smaller numbers 
of potential slots are considered first. Then for each train the slots are labelled
with trips: first slot first, last slot last. The trips are chosen according to their
index: greatest index first, smallest index last such that 'regular' trips are
considered first. For the first slot the 'virtual' initial trip is fixed and for
the other slots the 'real' trips are chosen first. Subsequently the maintenance 
tasks are labelled between trips starting with the ``no maintenance'' index ($-1$) 
first to avoid maintenance tasks if not required. Tree search and constraint processing, i.e. mainly filtering, was performed on an Intel(R) Xeon(R) CPU E5-2695 
v4 @ 2.10 GHz (in single core mode) running Ubuntu 20.04.2 and using OpenJDK 1.8.0,
which is the basis of  the  implemented CP approach. In particular, the
\emph{first-fail} heuristic only influences the order of variable selection for assignment during search. There are no branches cut in the search tree.

The QUBO models are evaluated on the D-Wave hybrid (classical+quantum) cloud system named LEAP, which integrates the 5760 qubit machine ``Advantage'', as well as on a classical computer employing tabu search. The classical computer used has 512 GB of RAM and 36 cores (72 with hyper-threading).

\begin{table}[htbp]
    \caption{Results of the empirical examination.
    }
    \label{tab:results}
    \setlength{\tabcolsep}{3pt} %% change for appropriate horizontal spacing
    \renewcommand{\arraystretch}{1.0} %% change for appropriate vertical spacing
    \centering
    \begin{tabular}{l|r|r|r|r|lr}
        \multirow{2}{*}{data set}& \multirow{2}{*}{\# qubits} & \# trips
        & \# trains &  empty 
        & \multirow{2}{*}{method} & \multirow{2}{*}{run-time} \\
        &  & alloc. & used & rides [km] & &  \\ \hline\hline

        \multirow{2}{*}{real-small}&  & 52 &  34 &   3769 
        & CP first&     $<$ 1 sec \\ 
        & & 52 & 34 &  3007 & CP improved &    1 sec \\    
        70 trips & \multirow{2}{*}{4662} & 52(51) & 35 &  1973 & LEAP & 1+2 min \\ 
        38 trains & &52(51) & 35 &  289 & tabu search & 1+2 min \\        
        \hline

        \multirow{2}{*}{real-50\%} & & 121 &  70 &
         8694  & CP first &     2 sec \\ 
        &  & 123 & 70 & 8092 & CP improved & 1.23 hours \\  
        141 trips & \multirow{2}{*}{20256} &123(120)  & 75 &  7158  & LEAP & 8+60 min \\ 
        75 trains & & 123(119) & 73 &  4231 & tabu search & 8+28 min \\        
        \hline 

        \multirow{2}{*}{real-75\%} & & 189 & 106 
        & 8310 & CP first & 7 sec \\ 
        &  & 190 & 106 & 8710 & CP improved & 10 sec \\ 
        212 trips & \multirow{2}{*}{47358} & 183(177)  & 112 &  11936 
        & LEAP & 0.5+1 hours \\
        112 trains & & 186(185) &  110 &  6550 & tabu search & 0.5+2.3 hours \\        
        \hline
        \multirow{2}{*}{real-100\%} & & 252 &  139 
        &   14841 & CP first & 11 sec \\  
        & & 252 & 139 &  14552 & CP improved & 17 sec \\  
        284 trips & \multirow{2}{*}{87069} & 243(240) & 150 &  17657 & LEAP 
        & 1.25+2 hours \\
        150 trains & & 246(239) & 146 & 8762 & tabu search & 1.25+5.5 hours 
    \end{tabular}
\end{table}

Table~\ref{tab:results} shows the results of the empirical examination of both approaches on the generated data sets.
For better alignment of the CP and QUBO approaches we are using 3~time slots per train.
We show different data sets with the number of available trips and trains, the 
numbers of the required qubits in the QUBO models, the numbers of allocated trips, the 
numbers of used trains, the total sum of kilometres for empty rides, the applied methods, 
and the required run-times for finding the solutions. Due to the fact that only for the 
largest considered data set, ``real-100\%'', the CP approach takes more than 1~second, 
i.e., 1.5 seconds, to establish the constraints and perform an initial propagation of 
these constraints, only the run-time for searching is shown in the table --- the time for 
problem generation is omitted. The first solutions and the improved solutions found so 
far (within hours), i.e. with the greatest number of allocated trips and smallest sum of 
empty driven kilometres, are shown. 
By definition these solutions satisfy the constraints (cf.~Sec.~\ref{sec:CP}) , i.e. the 
allocated trips can be performed and the limits of the maintenance types are respected. 

\begin{figure}[htbp]
\centering
 \includegraphics[width=\textwidth]{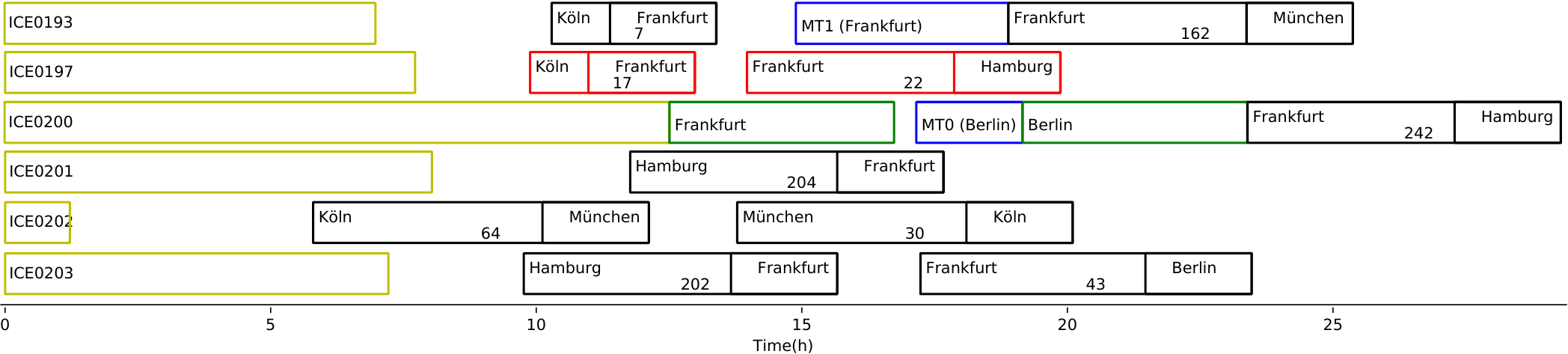}
 \caption{Extract from a QUBO-based solution, showing regular trips (black), maintenance 
 (blue), empty travel~(green), unavailability (yellow) and conflict with the maintenance 
 requirements (red).}
 \label{fig:results}
\end{figure}

For the QUBO approaches, i.e. LEAP and tabu search, two numbers for the allocated trips 
are given. The first one is the number provided by the solution, the second one, in 
parentesis, is lower and represents the number of trips after removing the trips which 
violate maintenance constraints. This can also be seen in Figure~\ref{fig:results}, 
where an extract of a QUBO solution is shown.
The time for computing the solution is split into pre-processing and the actual search 
time. Here, the main pre-processing task is the calculation of the QUBO matrix, which can 
be very time-consuming for a large number of qubits, but still polynomial.
In contrast to other quantum computing cloud providers such as IBM, in the case of D-Wave the time spent queuing was negligible, the time we reported almost equals the processing time.

\section{Discussion} \label{sec:discuss}
For CP, in all cases the number of allocated trips in the first solution (found within 
seconds) are rather good, i.e. it is not significantly increased while searching for 
better solutions. Spending hours searching for better solutions only results in either 
one or two additional allocated trips or in moderate reductions (2\% resp. 25\%) of empty
driven kilometres.
For the ``real-small'' subset only one improved solution was found rather quickly after 
1 second, however ongoing search for several days failed to find a better 
solution. 
For each examined data set, it was impossible to prove with the CP approach that one 
of the found solutions is optimal. 
We used 
%our own TODO: RESTORE
a
CP solver for the examinations that implements state-of-the-art filtering algorithms and techniques of modern CP solvers (cf.~\cite{schulteEfficientConstraintPropagation2008}) because there we were able to 
implement special filtering algorithms for the logical implications and 
the extended element constraints used in the CP model. 
%Our TODO: RESTORE
This CP solver was sufficient to show that a CP approach is currently able to outperform a Quantum / QUBO approach.

% The QUBO solutions are only slightly worse than the CP solutions, in terms of allocated 
% trips and number of trains used. On the other hand, the empty driven kilometres have the 
% lowest value for classical tabu search. 

The numerical results presented in Table 1 are of similar quality, as the QUBO solutions cover only up to 3~\% fewer trips than the CP solutions (up to 7~\% fewer trips when only the LEAP solutions are considered), while the empty driven kilometres are of the same order of magnitude. This way the QUBO approach is able to produce qualitatively similar results as the CP approach. Of course, the run times for both approaches differ widely. 
We noticed that the empty driven kilometres have the lowest value for classical tabu search.

The run times for the QUBO approaches are much 
longer than for the CP cases. However, the LEAP cloud hybrid solver from D-Wave shows in 
the online dashboard that the amount of time spent on the QPU never surpassed three 
seconds, even when the total time spent by LEAP was up to 2~hours.

\section{Conclusion and Future Work} \label{sec:concl}
We found once again that problem-specific constraints and heuristics are required to being 
able to handle realistic problem sizes. We were thus able to optimise the high-speed Intercity-Express (ICE) railway traffic in Germany that goes through 5 major cities. 
Surprisingly, the QUBO-based method was able to handle up to almost 100,000 qubits.
In general, the CP solution outperformed the QUBO solution except for the real small 
subset. Considering the current stage of the quantum annealer and of the hybrid solver 
from D-Wave, the results obtained on classical computers and using the quantum annealer 
are fairly comparable.
We plan to extend the amount of cities covered and handle more details of the real problem 
(e.g. the intermediary stations of the trips) towards a practice-ready prototype.
For a stable operation, it is also advantageous to optimise the time slots when trains are 
not in operation, which, ideally, are as long as possible, instead of scheduling many 
short break time slots. This increases the chance that one train can replace another 
cancelled train. 
Hopefully, in the near future, the performance of the quantum annealers will increase 
beyond the capacity of classical computing.

\printbibliography

\end{document}